\documentclass[runningheads]{llncs}
\usepackage{graphicx}
\usepackage{tabto}
\usepackage{amsfonts}
\usepackage{enumitem}
\usepackage{wrapfig}
\usepackage{rotating,graphicx}
\usepackage{mathtools}
\usepackage{fancyhdr}
\usepackage{tabularx, booktabs} 
\usepackage{multirow}
\usepackage{gensymb}
\usepackage{hhline}
\usepackage{hyperref} 
\usepackage{subcaption}
\usepackage{float}
\usepackage{cite}
\usepackage{url}
\usepackage{algorithm}
\usepackage{algorithmic}
\usepackage{array, makecell}
\usepackage{xcolor}
\usepackage{commath}
\usepackage{amsmath}
\usepackage{wrapfig}
\usepackage{hyperref}
\usepackage[title]{appendix}
\usepackage{hyperref}
\begin{document}
\title{A Baseline Approach for AutoImplant: the MICCAI 2020 Cranial Implant Design Challenge\thanks{\url{https://autoimplant.grand-challenge.org/.}}}
\author{Jianning Li\inst{1,2}\and
Antonio Pepe\inst{1,2}\and
Christina Gsaxner\inst{1,2,3} \and
%Dieter Schmalstieg \inst{1} \and
Gord von Campe \inst{4} \and 
Jan Egger \inst{1,2,3}
} 

\authorrunning{A Baseline Approach for AutoImplant.}
\institute{Institute of Computer Graphics and Vision, Graz University of Technology, Graz, Austria. \and
Computer Algorithms for Medicine Laboratory (Caf\'e-Lab), Graz, Austria. \and
Department of Oral and Maxillofacial Surgery, Medical University of Graz, Graz Austria. \and
Department of Neurosurgery, Medical University of Graz, Graz, Austria \\
\email{\{jianning.li,egger\}@icg.tugraz.at}}

\maketitle              % typeset the header of the contribution
\begin{abstract}
In this study, we present a baseline approach for AutoImplant (\url{https://autoimplant.grand-challenge.org/}) -- the cranial implant design challenge, which, as suggested by the organizers, can be formulated as a volumetric shape learning task. In this task, the defective skull, the complete skull and the cranial implant are represented as binary voxel grids. To accomplish this task, the implant can be either reconstructed directly from the defective skull or obtained by taking the difference between a defective skull and a complete skull. In the latter case, a complete skull has to be reconstructed given a defective skull, which defines a volumetric shape completion problem. Our baseline approach for this task is based on the former formulation, i.e., a deep neural network is trained to predict the implants directly from the defective skulls. The approach generates high-quality implants in two steps: First, an encoder-decoder network learns a coarse representation of the implant from down-sampled, defective skulls; The coarse implant is only used to generate the bounding box of the defected region in the original high-resolution skull. Second, another encoder-decoder network is trained to generate a fine implant from the bounded area. On the test set, the proposed approach achieves an average dice similarity score (DSC) of 0.8555 and Hausdorff distance (HD) of 5.1825 mm. The code is publicly available at \url{https://github.com/Jianningli/autoimplant}.        

\end{abstract}
\section{Introduction}
\begin{figure}
     \centering
        \includegraphics[width=\textwidth]{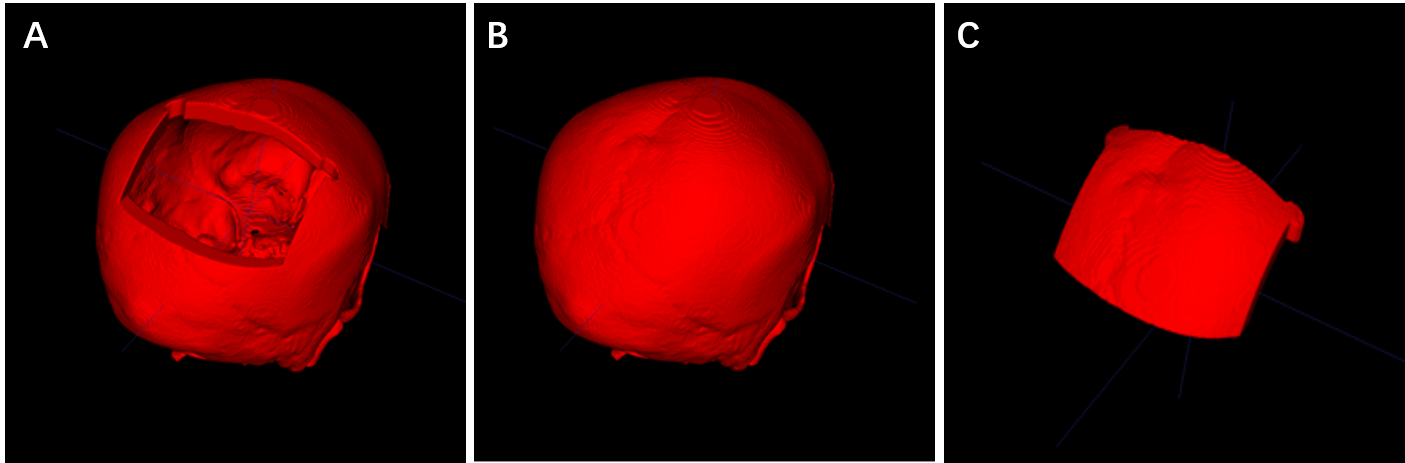}
     \caption{Illustration of a defective skull (A), complete skull (B) and the implant (C). The defective skull is created by removing a bony part (i.e., the implant) from the complete skull.}
     \label{fig:fig1}
\end{figure}
In current clinical practice, the process of cranial implant design and manufacturing is performed externally by a third-party supplier. The process usually involves costly commercial software and highly-trained professional users \cite{casestudy}. A fully automatic, low-cost and in-Operation Room (in-OR) design and manufacturing of cranial implants can bring significant benefits and improvements to the current clinical workflow for cranioplasty \cite{li2020online}.
Previous work has seen the development of freely available CAD tools for cranial implant design \cite{Gallinproceedings,Chenarticle, Marzolaarticle,Janarticle}, whereas these approaches are still time-consuming and require human interaction. These approaches tend to exploit the geometric information of skull shape. For example, one of the most popular techniques used in these approaches is to find the symmetry plane of the skull and fill the defected region by mirroring \cite{Angeloarticle}. Considering that human skulls are not strictly symmetric, mirroring is not an optimal solution. \\
The AutoImplant Challenge \cite{autoimplant} aims at solving the problem of automatic cranial implant design in a data-driven manner, without relying explicitly on geometric shape priors of human skulls. As suggested by the organizers, cranial implant design can be formulated as a volumetric shape learning task, where the shape of the implant can be learnt directly or indirectly from the shape of a defective skull \cite{li2020online}. On the one hand, the shape of the implant can be directly learnt from a defective skull. On the other hand, by learning to fill the defected region on a defective skull, a completed skull can be produced. The implant can then be obtained indirectly by taking the difference between the completed skull and the defective skull. In this sense, cranial implant design is being formulated as a shape completion problem \cite{Dai2016ShapeCU, Sung2015DatadrivenSP, Sarmad2019RLGANNetAR,Stutz2018Learning3S,Han2017HighResolutionSC}. A relevant study was conducted by Morais et al. \cite{Morais2019}, where an encoder-decoder network is used to predict a complete skull from a defective skull. However, the study deals with very coarse skulls of low dimensionality ($30^3$, $60^3$ and $120^3$) extracted from MRI data, whereas, in practice, the common imaging modality used for head scans acquisition is computed tomography (CT), with a typical resolution of $512 \times 512 \times Z$. In this study, we primarily elaborate on the former formulation, i.e., given a defective skull shape, we directly predict the shape of the implant, which is a challenging task as the implant has to be congruent with the defective skull in terms of shape, bone thickness and boundaries of the defected region\cite{li2020online}. \\
The defective skulls used in our datasets are created artificially out of complete skulls. By doing so, we have a ground truth for supervised training for either of the two above mentioned problem formulations. For direct implant generation, the ground truth is the implant, which is the region removed from a complete skull. For skull shape completion, the ground truth is the original complete skull. The input of either formulation is the defective skull. Real surgical defects from a craniotomy surgery are usually more complex and irregular than the artificial defects. However, we expect that the deep learning networks trained on artificial defects can be generalized to the real surgical defects in craniotomy, which requires that the networks should be robust as to the shape, position and size of the defects.

\section{Dataset}
The 200 skull datasets (100 for training and 100 for testing) are selected from QC 500 (\url{http://headctstudy.qure.ai/dataset}), which is a public collection and contains 491 anonymized head CT scans in DICOM format. Considering that the datasets are acquired from patients with various head pathologies, we discarded the scans that present a severe skull deformity or damage. Lower-quality scans (e.g., $z$-spacing above 1 millimeter) were also discarded. The dimension of these skulls is $512 \times 512 \times Z$, where $Z$ is the number of axial slices. For ease of use, the selected DICOM scans were converted to the NRRD format. To extract the binary skull data, a fixed threshold (Hounsfield units values from 150 to maximum) was applied to the CT scans. As the thresholding also preserves the CT table head holder, which has a similar density to the bony structures, we used 3D connected component analysis to automatically remove this undesired component. The last step is to generate an artificial surgical defect on each skull, which was accomplished by removing a bony structure from the skull. The data processing step is summarized as follows:       
\begin{enumerate}
    \item \textbf{DICOM Selection:} 200 High quality DICOM files were selected.
    \item \textbf{NRRD Conversion:} DICOM files were converted into NRRD format.
    \item \textbf{Skull Extraction:} Skulls were extracted using thresholding (150 HU-Max).
    \item \textbf{CT Table Removal:} The CT table head holders were removed.
    \item \textbf{Hole Injection:} On each skull, an artificial surgical defect was injected.
\end{enumerate}
\autoref{fig:fig1} shows a defective skull, the corresponding original skull and the implant (i.e., the removed part) in the training set. The skull defects shown in \autoref{fig:fig1} (A) are representative of those of the 100 training datasets and 100 test datasets. The final evaluation of implant generation algorithms will be based on the 100 test datasets. However, we create another test set containing 10 defective skulls, which have completely different defects from those in the 100 training and test data, in terms of shape, size and position of the defects. Currently, the additional 10 test data are not involved in the quantitative evaluation of the proposed approach. 
\begin{figure}[t]
     \centering
        \includegraphics[width=\textwidth]{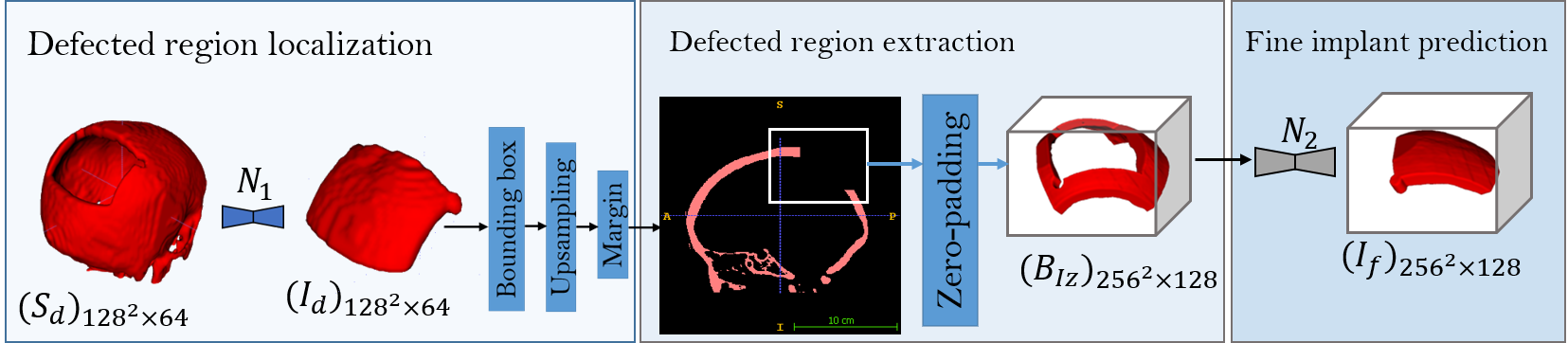}
     \caption{The workflow of the proposed approach.}
     \label{fig:overview}
\end{figure}
\section{Method}
\begin{figure}
     \centering
        \includegraphics[width=\textwidth]{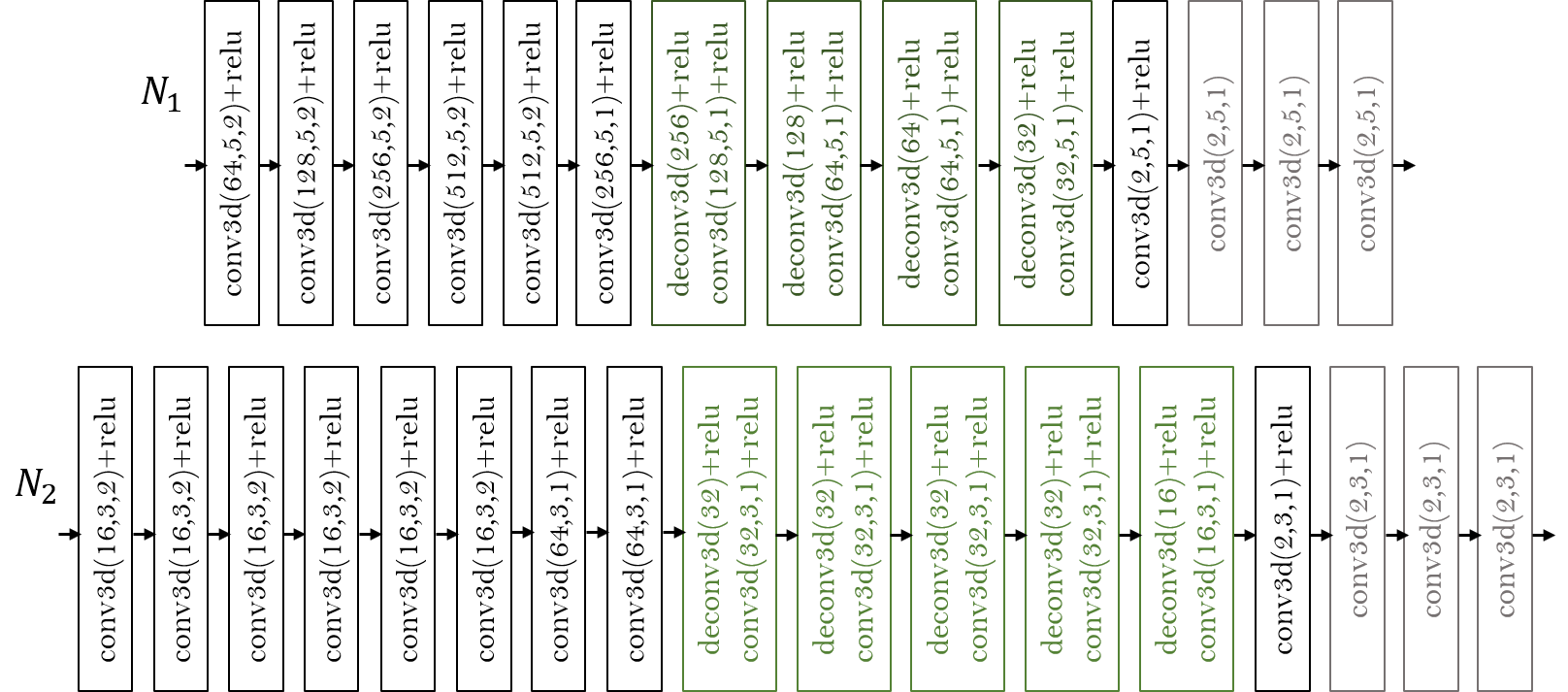}
     \caption{Detailed architecture of $\mathbf{N}_1$ and $\mathbf{N}_2$. Both $\mathbf{N}_1$ and $\mathbf{N}_2$ use two-strided convolution for down-sampling and de-convolution for up-sampling.}
     \label{fig:network}
\end{figure}

The proposed implant generation scheme is illustrated in \autoref{fig:overview} and consists of three steps. $\mathbf{First}$, an encoder-decoder network $\mathbf{N}_1$ learns to infer a coarse implant $(\mathbf{I})_{128^2\times 64}$ directly from a coarse defective skull $(\mathbf{S}_d)_{128^2\times 64}$, which is downsampled from the original high-resolution defective skull $(\mathbf{S}_d)_{512^2\times Z}$. This allows to contain the requirements of GPU memory:
\begin{equation}
(\mathbf{I}_c)_{128^2\times 64}\overset{\mathbf{N}_1}{\leftarrow}(\mathbf{S}_d)_{128^2\times 64} .
\label{eq:coarse_implant}
\end{equation}
$\mathbf{Second}$, we calculate the bounding box $(\mathbf{B}_{I})_{X_B \times Y_B \times 128}$ of the coarse implant predicted by $\mathbf{N}_1$, which is then used to localize the defected region on the high-resolution defective skull. $X_B$ and $Y_B$ are dimensions of the bounding box in $x/y$ volume axis. In the $z$ axis, we fix the dimension to 128 (the maximum $z$ dimension of the defected area in the challenge dataset is smaller than 128). Considering that the bounding box tightly encloses the defected region in the $x/y$ axis, a margin $\mathbf{m}$ is used to keep some surrounding information around the defected region. In order to get a fixed bounding box dimension $(\mathbf{B}_{Iz})_{256^2 \times 128}$, zero-padding is applied. $\mathbf{Third}$, a second encoder-decoder network $\mathbf{N}_2$ learns to infer the fine implants $(\mathbf{I}_f)_{256^2 \times 128}$ from the bounded region of the high-resolution defective skulls:
\begin{equation}
(\mathbf{I}_f)_{256^2 \times 128}\overset{\mathbf{N}_2}{\leftarrow}(\mathbf{B}_{Iz})_{256^2 \times 128}
\label{eq:fine_implant}
\end{equation}
The detailed architecture of $\mathbf{N}_1$ and $\mathbf{N}_2$ is shown in \autoref{fig:network}. As the input size of $\mathbf{N}_2$ is larger than that of $\mathbf{N}_1$, the complexity of $\mathbf{N}_2$ has to be significantly reduced compared to $\mathbf{N}_1$ in order to get the network running on the limited GPU memory. 
%Equation ~\eqref{eq:fine_implant}
In particular, the kernel size of all convolutional layers in $\mathbf{N}_1$ is five, whereas the kernel size for $\mathbf{N}_2$ is only three. The number of feature maps of each layer for $\mathbf{N}_2$ is also significantly reduced, resulting in a total of $0.6538$ million trainable parameters, compared to $82.0766$ million parameters for $\mathbf{N}_1$ .
\autoref{fig:overview} shows the $input/output$ of $\mathbf{N}_1$ and $\mathbf{N}_2$. $\mathbf{N}_1$ takes as input a downsampled defective skull and produces a coarse implant prediction. $\mathbf{N}_2$ takes as input a zero-padded version of the high-resolution defected area delimited by the bounding box $(\mathbf{B}_{Iz})_{256^2 \times 128}$ and produces a prediction of the corresponding fine implant.
%\begin{figure}
     %\centering
      %  \includegraphics[width=\textwidth]{figures/bbox.png}
     %\caption{An illustration of the $input/output$ of $\mathbf{N}_1$ (A) and $\mathbf{N}_2$ %(B).}
%     \label{fig:bbox}
%\end{figure} 
The bounding box and the amount of zero-padding are calculated as follows:
\paragraph{Bounding Box} The bounding box of an implant is calculated by finding the coordinates of the first and last non-zero values in the projected $x/y$ plane of the image volume $(\mathbf{V})_{512^2 \times Z}$. The predicted coarse implant $(\mathbf{I})_{128^2\times 64}$ is first upsampled to its original dimension $512^2 \times Z$ using a (order two) spline interpolation before the bounding box is calculated. The bounding box $(\mathbf{B}_{I})_{X_B \times Y_B \times 128}$ tightly encloses the defected region in the original high-resolution skull. We apply an additional margin $\mathbf{m}=20$ in the $x/y$ direction to also enclose a portion of the surrounding skull, information which facilitates the learning task. The $x/y$ dimension of the resulting bounding box becomes: $X_B+2m$ and $Y_B+2m$.           
\paragraph{Zero-padding} As the dimension of each bounding box is different, we apply zero-padding on the bounding boxes to obtain inputs with a fixed dimension $(\mathbf{B}_{Iz})_{256^2 \times 128}$ for the deep neural network. Zero-padding is done by moving the bounding box to the middle of an all-zero volume of dimension $256^2 \times 128$.  
\section{Experiments and Results}
\begin{figure}
     \centering
        \includegraphics[width=\textwidth]{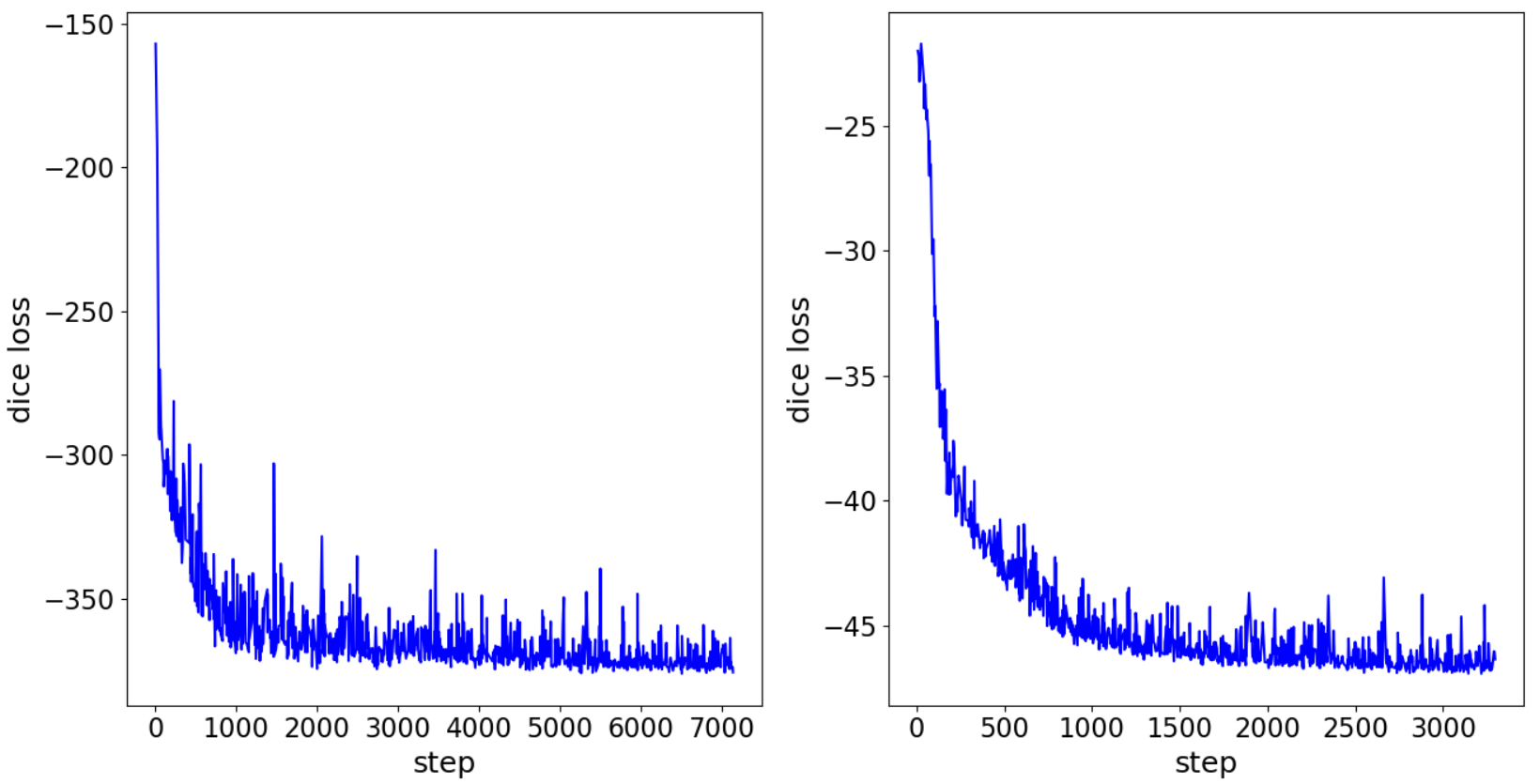}
     \caption{The $step/loss$ curve of $\mathbf{N}_1$ (left) and $\mathbf{N}_2$ (right) during training.}
     \label{fig:training_loss}
\end{figure}

$\mathbf{N}_1$ and $\mathbf{N}_2$ were consecutively trained on a machine equipped with one GPU NVIDIA GeForce GTX 1070 Ti, which presents a limited GPU memory of 8 GB. First, $\mathbf{N}_1$ was trained on downsampled defective skulls. Once the training of $\mathbf{N}_1$ was completed, we used $\mathbf{N}_1$ to produce coarse implants on the training set. Then, the coarse implants were upsampled to their original size of each corresponding training sample. Second, the upsampled implants were used to calculate the bounding box of the defected region on the high-resolution defective skulls. The bounding box, extended by a margin of $2\times m$ to include a portion of skull, was used to train $\mathbf{N}_2$. The networks were trained on the 100 data pairs provided by the AutoImplant challenge, without using any additional dataset or defect shapes for data augmentation. Additionally, it needs to be considered that the performance of $\mathbf{N}_2$ depends on the accuracy, or failure rate, of $\mathbf{N}_1$. In both cases the batch size was set to one. We employed dice loss as a loss function, which measures the shape similarity between a predicted implant and its corresponding ground truth implant. \autoref{fig:training_loss} shows the $step/loss$ curve during training. Fluctuations in $\mathbf{N}_1$ are due to the small batch size. In contrast, $\mathbf{N}_2$ shows a smoother curve despite the batch size.    
Shape similarity between the predicted implant and the ground truth is quantitatively evaluated using the Dice similarity score (DSC), the symmetric Hausdorff distance (HD) and the reconstruction error (RE). The RE for each test case is defined as the false voxel prediction rate as in \cite{Morais2019}:
\begin{equation}
RE = \frac{\sum|\mathbf{P}-\mathbf{G}|}{N} \label{eq:re}
\end{equation}
\begin{figure}[ht]
     \centering
        \includegraphics[width=\textwidth]{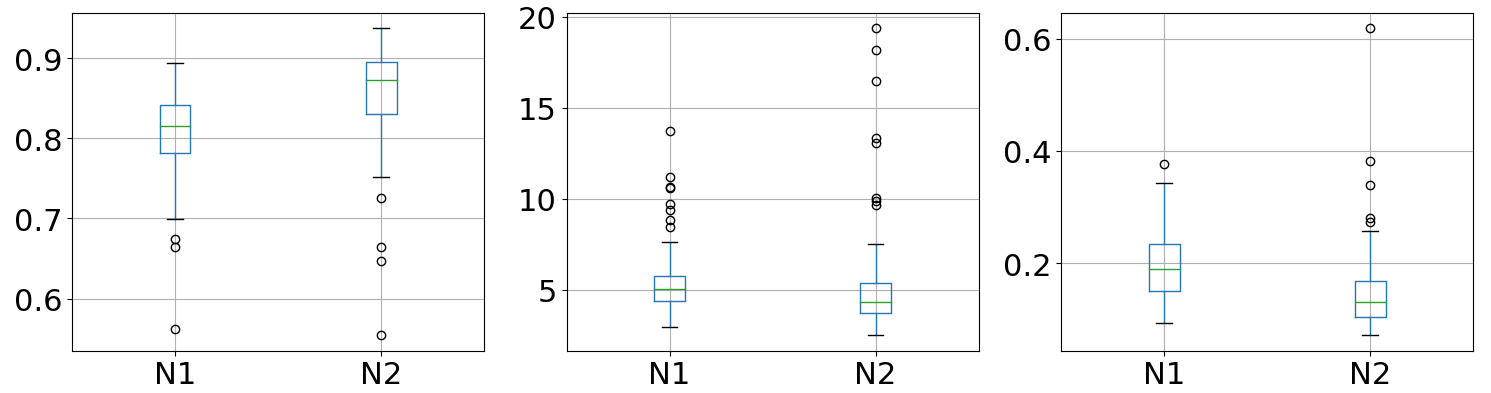}
     \caption{Boxplots of DSC (left), HD (middle) and RE (right) of the 100 test cases.}
     \label{fig:boxplot}
\end{figure}
$(\mathbf{P})_{512^2 \times Z}$ and $(\mathbf{G})_{512^2 \times Z}$ are the fine implant produced by $\mathbf{N}_2$ and its corresponding ground truth, respectively. $\sum|\mathbf{P}-\mathbf{G}|$ represents the total number of voxels in $\mathbf{P}$ that are different from $\mathbf{G}$. $N=512^2 \times Z$ is the total number of voxels in the volume. Note that, in order to calculate the metrics -- DSC, HD, and RE -- against the ground truth, which has a $512^2 \times Z$ dimension, the corresponding inverse process of zero-padding and bounding box was applied to the prediction $(\mathbf{I}_f)_{256^2 \times 128}$ from $\mathbf{N}_2$ so that the prediction $(\mathbf{P})_{512^2 \times Z}$ was of the same dimension as that of the ground truth $(\mathbf{G})_{512^2 \times Z}$. Similarly, to calculate the metrics for $\mathbf{N}_1$, the coarse implants $(\mathbf{I}_c)_{128^2\times 64}$ were upsampled to their corresponding original dimensions $512^2 \times Z$ using interpolation. To provide the HD in millimeters (mm), we considered the actual image spacing of each test case, which is provided in the header of the NRRD files. \autoref{tab:quantitative_results} shows the mean value of DSC, HD, and RE on the 100 test cases. The corresponding boxplot is shown in \autoref{fig:boxplot}. 
\begin{table}
\centering
\caption{Quantitative Results}
\begin{tabular}[t]{lccc}
\toprule
\hspace{0.6cm} &DSC  \hspace{1cm} & HD (mm) \hspace{1cm} & RE (\%)\\
\midrule
$\mathbf{N}_1$ \hspace{0.6cm} &0.8097 \hspace{1cm} & 5.4404 \hspace{1cm} & 0.20 \\
$\mathbf{N}_2$ \hspace{0.6cm} &0.8555 \hspace{1cm} &5.1825 \hspace{1cm} &0.15 \\
\bottomrule
\end{tabular}
\label{tab:quantitative_results}
\end{table}%
\autoref{fig:results} gives an illustration of the automatic implant generation results in 3D for five test cases (A-E). We can see that the implants from $\mathbf{N}_1$ are coarse (second column), lacking geometric details compared to the ground truth (fourth column). The reason is that the implants are learnt from downsampled skulls (($\mathbf{S}_d)_{128^2\times 64}$), which are already deviating from the original high-resolution skulls. In comparison, $\mathbf{N}_2$ produces fine, high-quality implants (third column), which are close to the ground truth, as $\mathbf{N}_2$ learns directly from high-resolution skull shapes. We also empirically noticed how $\mathbf{N}_2$ captures highly intricate details such as the smoothness of the implant surface and the details of the small roundish corners of the implants, which are not well preserved in the coarse implants generated by $\mathbf{N}_1$. Furthermore, (A'-E') show how the fine implants generated by $\mathbf{N}_2$ match with the defected region on the defective skulls in terms of shape and bone thickness. \autoref{fig:zoom_in} (A) shows a zooming in of the coarse implant (left) generated by $\mathbf{N}_1$, the fine implant (middle) generated by $\mathbf{N}_2$ and the ground truth (right). \autoref{fig:zoom_in} (B) show how shape of the predicted implant (red) matches with that of the ground truth (white) in 2D axial, sagittal and coronal view.     
\begin{figure}
     \centering
        \includegraphics[width=0.9\textwidth]{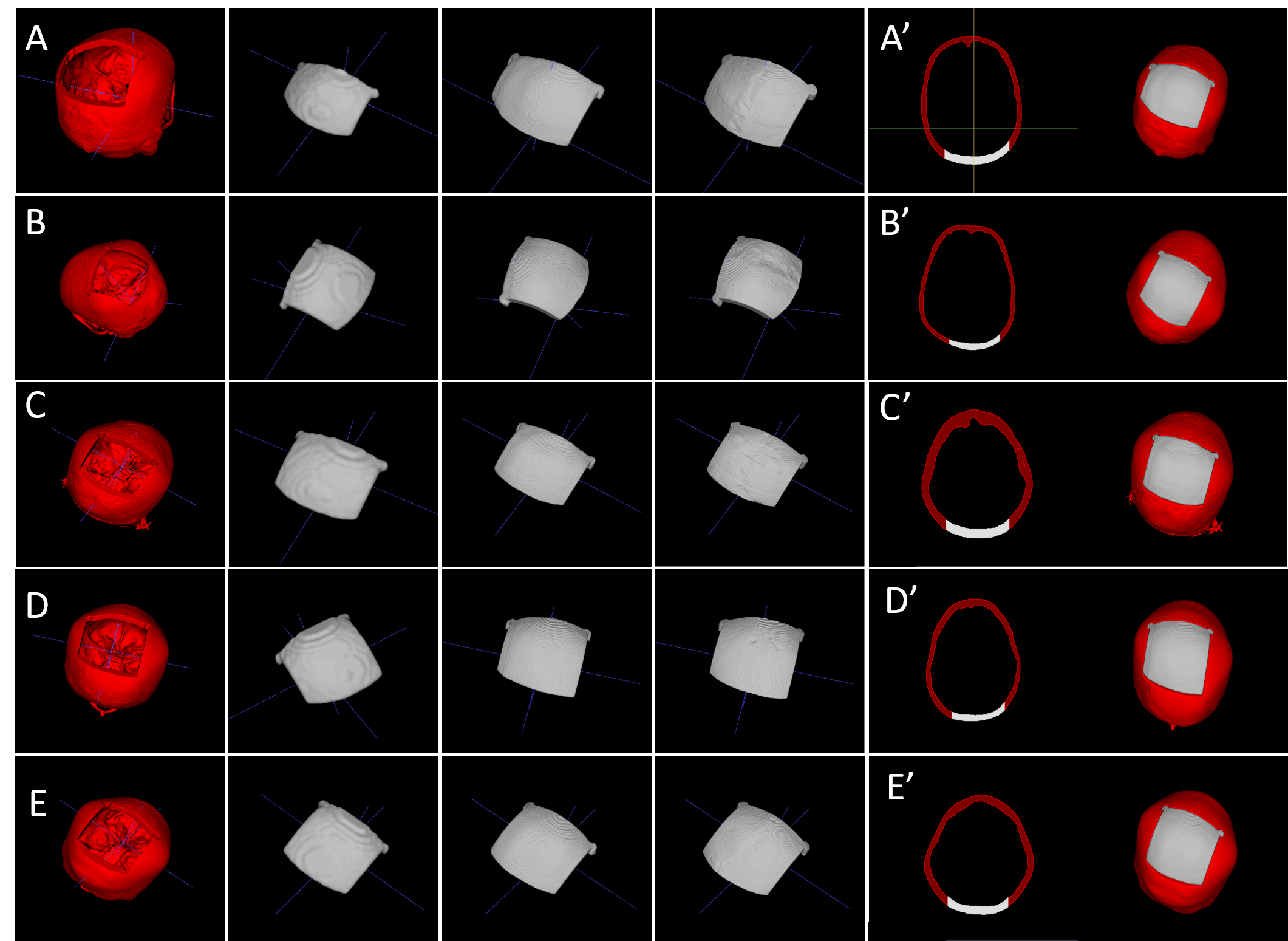}
     \caption{(A)-(E) implant prediction results on five evaluation cases. From left to right: the input defective skulls; the coarse implants from $\mathbf{N}_1$; the fine implant predictions from $\mathbf{N}_2$; the ground truth. (A')-(E') overlay of the implants from $\mathbf{N}_2$ on the defective skulls in 2D axial view (fifth column) and in 3D (sixth column). To differentiate them, different colors are used for the implants (gray) and skulls (red).}
     \label{fig:results}
\end{figure}
\begin{figure}
     \centering
        \includegraphics[width=0.8\textwidth]{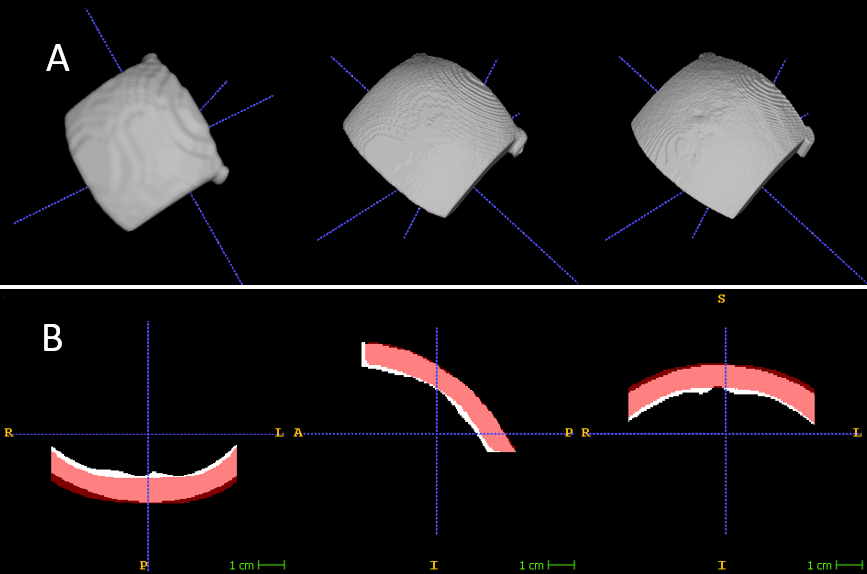}
     \caption{(A):A zooming in of the coarse implant (left), fine implant (middle) and the ground truth (right). (B): how the shape of the predicted implant matches that of the ground truth in 2D.}
     \label{fig:zoom_in}
\end{figure}
\section{Discussion}
As introduced, the missing portion of an incomplete shape can be estimated in two alternative ways. The first is to $directly$ reconstruct the missing part and the second is to predict a completed shape first and then generate the missing part $indirectly$ by taking the difference between the completed and defective shape. Learning the missing part directly is prone to two types of overfitting. First, the network $memorized$ the shape of skulls during training and always gives the same output given the same skull even if the defect on the skull has been changed. Second, the network cannot be generalized to hole shapes that are different from the training set. Both types of overfitting are undesirable during the shape learning process. On the 10 additional test sets, where the injected defects are completely different from those in the training and test set, the proposed direct implant generation approach fails. As a comparison, we trained $\mathbf{N}_1$ for skull shape completion and tested the network on both the 100 test sets and the additional 10 test sets. Some selected results are shown in Appendix~\ref{appendix:volumetric_shape_completion} and Appendix~\ref{appendix:robustness_testing}. We can see that $\mathbf{N}_1$ can work well on both the defective skulls from the 100 test sets and skulls from the additional 10 test sets, even if $\mathbf{N}_1$ was trained only on the 100 training set provided by the challenge without using data augmentation (i.e., creating own defective skulls with varied defects). This shows that the skull shape completion network can be well generalized to different hole shapes, positions and sizes, which is a desirable property for cranial implant design as the defects caused in craniotomy are varied depending on the intracranial pathology to be operated on for each particular patient.

\section{Conclusion and Future Improvement}
The contribution of this study is threefold. $\mathbf{First}$, we demonstrated that a fully data-driven approach without using geometric priors can be effective in high-resolution volumetric shape learning. An encoder-decoder network $\mathbf{N}_1$ can directly learn to reconstruct the missing part from a defective skull. $\mathbf{Second}$, we show that an encoder-decoder network $\mathbf{N}_2$ does not need to see the entire skull shape to predict the missing part. Instead, the learning can be based only on the defected region with limited surrounding shape information. $\mathbf{Third}$, we provide a baseline approach for the AutoImplant Challenge, which is in essence a volumetric shape learning task. To contain the GPU memory requirements, $\mathbf{N}_2$ has a much lower complexity than $\mathbf{N}_1$, which is a potential performance bottleneck. Increasing the network complexity (e.g., number of trainable parameters) could, in future analyses, lead to an increased performance of the network and therefore even more accurate cranial implants.

\section*{Acknowledgment}
This work received the support of CAMed - Clinical additive manufacturing for medical applications (COMET K-Project 871132), which is funded by the Austrian Federal Ministry of Transport, Innovation and Technology (BMVIT), and the Austrian Federal Ministry for Digital and Economic Affairs (BMDW), and the Styrian Business Promotion Agency (SFG). Further, this work received funding from the Austrian Science Fund (FWF) KLI 678-B31 (enFaced - Virtual and Augmented Reality Training and Navigation Module for 3D-Printed Facial Defect Reconstructions) and the TU Graz Lead Project (Mechanics, Modeling and Simulation of Aortic Dissection). Moreover, we want to point out to our medical online framework Studierfenster (\url{www.studierfenster.at}), where an automatic cranial implant design system has been incorporated. Finally, we thank the creators of the QC 500 dataset (\url{http://headctstudy.qure.ai/dataset}). 

%\newpage
\bibliographystyle{splncs04}
\bibliography{references}

% ---- Appendices ----
%\newpage
\begin{subappendices}
\renewcommand{\thesection}{\Alph{section}}%
% or try \arabic{section}

\section{Volumetric Shape Completion}
\label{appendix:volumetric_shape_completion}
\begin{figure}[H]
     \centering
        \includegraphics[width=\textwidth]{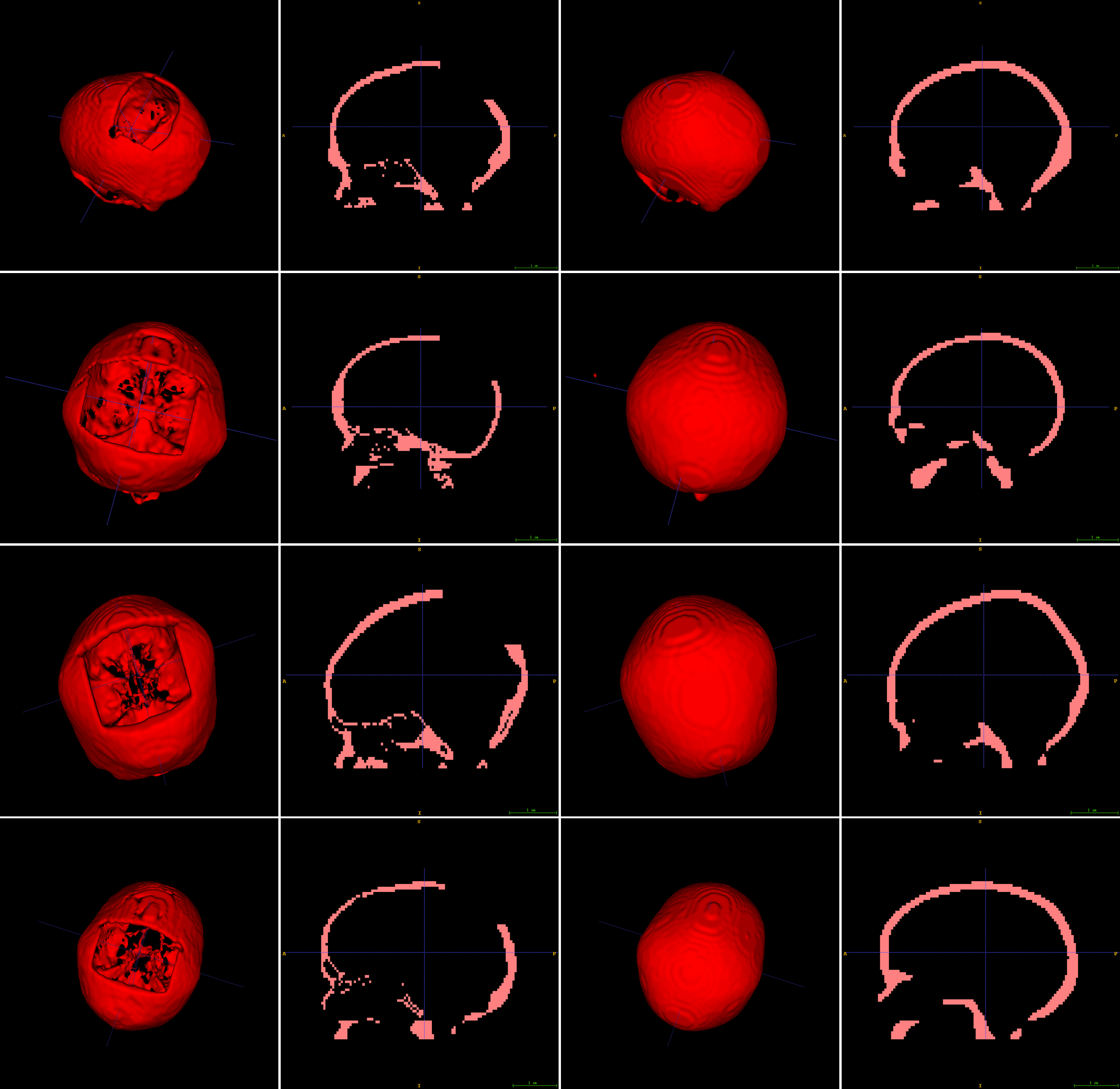}
     \caption{Skull shape completion results from $\mathbf{N}_1$, which is trained on defective skulls with the complete skulls as ground truth. Different from the direct implant prediction approach elaborated in the main text, which uses the implants as ground truth, for skull shape completion, $\mathbf{N}_1$ takes as input a defective skull (first and second column) and gives as output a completed skull (third and fourth column).}
     \label{fig:completion}
\end{figure}
\section{Robustness Testing of the Skull Shape Completion Network}
\label{appendix:robustness_testing}
\begin{figure}
     \centering
        \includegraphics[width=\textwidth]{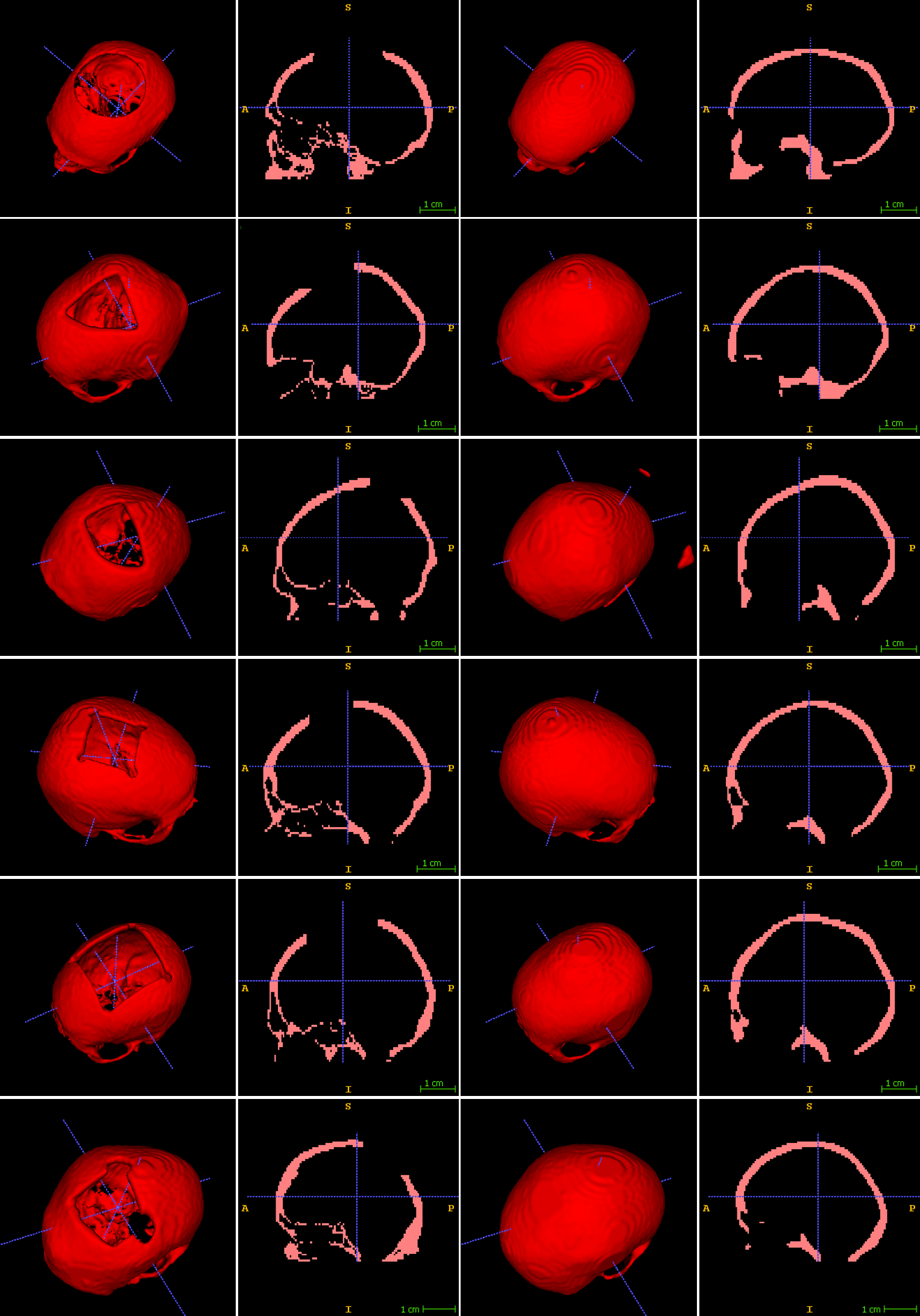}
     \caption{Testing the skull shape completion network using various defects (i.e., different size, shape, position of the defects). For skull shape completion, $\mathbf{N}_1$ is trained only on the 100 training dataset provided by the challenge, without using external datasets or generating own defective skulls for data augmentation. \autoref{fig:fig1} (A) shows a typical skull defect in the training set. The input in 3D (first column) and 2D (second column). The output in 3D (third column) and 2D (fourth column).}
     \label{fig:robustness}
\end{figure}

\autoref{fig:robustness} shows the skull shape completion results on an additional test set, where the skull defects are completely different from those in the training set.
\end{subappendices}
\end{document}